# Auto Deep Compression by Reinforcement Learning Based Actor-Critic Structure

**Hamed Hakkak**[1]

**Abstract:**

Model-based compression is an effective, facilitating, and expanded model of neural network models with limited computing and low power. However, conventional models of compression techniques utilize crafted features [2,3,12] and explore specialized areas for exploration and design of large spaces in terms of size, speed, and accuracy, which usually have returns Less and time is up. This paper will effectively analyze deep auto compression (ADC) and reinforcement learning strength in an effective sample and space design, and improve the compression quality of the model. The results of compression of the advanced model are obtained without any human effort and in a completely automated way. With a 4-fold reduction in FLOP, the accuracy of 2.8% is higher than the manual compression model for VGG-16 in ImageNet.

**key words:** Rl algorithms, Image compression,

## 1. Introduction:

Many real-world applications such as robotics, automatic machines, neural network convolution are limited by precise budgeting and timing. Many programs focus on existing compression networks. [2-4] Nevertheless, all of them perform a manual compression for each layer in a ratio, which requires human expertise and limited search space.

Self-development, without any skill, makes RL a powerful tool and can be implemented in many ways, but in cases where large dimensions of the data and the environment are non-constant, its use is low [47]. DL's ability to learn complex patterns sometimes also requires a misleading classification [48]. In recent years, RL algorithms have been successfully integrated with deep NN, which have created new learning strategies. This integration has led to the approximation of RL functions using deep NN architecture or deep NN training using RL [46]. It is suggested to use enhanced learning for automated compression, which does not require human expertise. Due to the complexity of the model when compressing a network by a human resource specialist, choosing design based on sparsity and for different layers should be done. The accuracy of the compressed model is dependent on sparsity and requires a precise action space. Using the Deep Deterministic Policy Gradient (DDPG), the control of continuous compression ratio is offered instead of searching in a discrete space. In order to reduce the duration of training that is created by the learning boost, it is suggested

---

[1] PhD student, Department of Biomedical Engineering, Mashhad Branch, Islamic Azad University, Mashhad, Iran.

that the agent reward be adjusted to the model's performance approximation, without penalty.

Actar-critic based on policy gradient (PG-based AC approach) is widely studied to examine learning control problems. In order to increase the efficiency of predictive learning data in critic AC based on PG, studies have been conducted on how to use recursive least squares scheduling algorithms (RLS-TDs) to evaluate policies in recent years. In such a situation, the critic RLS-TD evaluates an unknown policy produced by a series of actors, but none of the policies produced by the current Actor have been produced. Therefore, the AC framework with the critic RLS-TD, becoming its fixed point of learning optimality, can not be verified. To solve this problem, a new AC framework called the Crypt Crypt PG (CIPG) is presented, which learns the current value-mode policy in an on-policy and improves the gradient to improve the rebound process. During each replication, the CIPG maintains policy parameters and establishes a proven policy by the critic RLS-TD. The analysis expands the previous PG convergence with a functional approximation to the critic RLS-TD. The simulation results show that the term "2L regulation" in CIPG critic and during the process has a good learning experience and CIPG has a better performance in learning and convergence faster than AC learning control methods [49].

Specifically, the DDPG agent processes the network in a layered manner. For each t-level function, the agent embeds the st layer and displays it with a ratio, accurate compression. After the layer is compressed, the agent is passed to the next $t + 1$ layer. In all compact layers, agent reward accurately calculates relevant information and computes total compression. In the end, the best exploration model has a short duration (1/10 epochs of initial training) to achieve the best performance [1].

History is related to applications such as AI mobile phones, automated cars and adverts that require accurate timing. This layer is related to cloud-based applications, which are of great importance to quality rather than time; therefore, after compression, there is no problem in the precision of work. To achieve the limited compression of FLOPs, the action space is limited, so that the compressed models by the agent are always trying to find the best result. For high-precision compression, the RFLOP reward brings new bridges, which combine precision and computation and precisely determine the amount of compression.

This algorithm is evaluated with ResNet [5], VGG [6] and [7] MobileNet on both CIFAR-10 and ImageNet. A series of experiments show that better performance has been achieved through manual exploration policies. By reducing the 4-fold FLOP, the VGG-16 channel is raised to the top with an accuracy of 2.7% [8] and the pruning channel to the top of the top with a precision of 0.3% [3]. Decreasing more FLOPs from 1.0 MobileNet [7] 2 Equals by 68.8% accuracy is given for Vertex 1, which is better for Pareto than 0.75 for MobileNet. At $1.49 \times$ pixels on the Titan Xp and 1.65x on the Android smartphone.

Deep models that were used before ImageNet training were designed to effectively replace manual features and perform tasks such as dimension recognition and semantic segmentation. This method is used to accelerate the deep VGG-16 model and to detect the Fast R-CNN object [8].

## 2. related works

CNN compression and acceleration Since the damage to the desired brain [9], extensive work has been done on compression CNN [2,9,10,11]. Measurements [12-14] and specific implementation [15-18], in essence, increase the CNN speed. The tensor muscle estimation [19-22] divides the weight into several pieces, and [23,24] accelerates the complete layers attached to the SVD. [4] Divides a layer to $1 \times 3$, $3 \times 1$, and [8] one layer to $3 \times 3$ and $1 \times 1$. The pruning channel [25-28] removes deviant channels on feature mappings. However, a common problem is determining the sparsity ratio for each layer.

Search for Neural Architecture. Many search models with enhanced learning or genetic algorithms are widely used to improve CNN [29-31]. [32] For searching, portable network blocks are suggested. NASNets has been designed more than manual architecture and [5.33] has been used in ImageNet [34]. Exploration has been made faster by changing the network [35]. Integrating learning boost leads to the natural compression of the model. At the same time, N2N [36] aims to search for a compact model with learning to strengthen. However, their mode of space is not comparable with ImageNet. The method is based on the need for RNN to generate policies, without the need for reward training, constant continuous factor control, auto-assured compressed compression, and limited FLOP compression options.

The search for nerve architecture has some implications for program synthesis and induction programming, the idea of searching for a program from a sample (Summers, 1977; Biermann, 1978). In learning the machine, the possible induction of the program has been successfully used in many fields, such as learning to solve Q and A simple (Liang et al., 2010; Neelakantan et al., 2015; Andreas et al., 2016), sorting out A list of numbers (Reed & de Freitas, 2015) and learning with very few examples (Lake et al., 2015).

### 2.1. Methods used in this field

Neural network exists in both compression and compression calculations, and is limited to expansion in embedded systems with hardware resources. To address this constraint, "deep compression" has a three-tier line: pruning, trained quantification, and Huffman coding. This network method only works by learning important communications. Next, the weight is measured in order to share, finally, the Huffman coding is used. After the first two steps, the grid is reset to measure the remaining fittings and centroids. Pruning reduces the number of connections by $9 \times 13 \times$. Quantization then reduces the number of bits per connection from

32 to 5. The VGG-16 size method has been reduced 49 times from 552 MB to 11.3 MB, which is without loss of accuracy [2].

Although the latest high-end processor and powerful GPUs are available, deeper cannulation neural networks (CNNs) are challenging for complex tasks such as image naming on mobile devices. To launch deep CNNs on mobile devices, a simple and effective plan for general CNN compression has been proposed, which includes three steps: (1) the selection of rank with the division of the various Bayesian matrices, (2) Tucker decomposition The tensor core and (3) the regeneration setup of accumulated loss of precision, which each step can be easily accomplished using the publicly available tool. The effectiveness of the proposed design is demonstrated with the testing of various CNNs (AlexNet, VGGS, GoogLeNet and VGG-16) on the smartphone. Reducing costs in model size, runtime, and energy consumption [20].

Deep Convulsion Neural Networks (CNN) have become a very promising way to detect an object, the number of times and the results of breaking the record for categorizing image and object recognition in recent years. However, CNN is very deep, it contains many layers with millions of parameters, and it stores a very large network model. The use of deep CNNs prohibits the limitation of hardware resources, especially mobile phones or other embedding devices. In this paper, the investigation of quantization methods of vector information theory for compression of CNN parameters is proposed with the storage problem of the model. Using the k-means clustering to weigh or direct quantization, we can achieve a very good balance between model size and accuracy of detection. To classify 1000 categories in the ImageNet Challenge, it is able to achieve a 16-24-fold network compression with only 1% loss of classification accuracy using CNN [21].

Neural network architectures are wider and deeper, and continue to be the most advanced mode for computer vision tasks, with the adoption of these networks with hardware and speed limitations. Conventional compression methods are used to solve this problem by manually modifying the architecture or by using predefined discoveries. Because the space of reduced architectures is very large, the change in the architecture of a deep neural network is a tough task. In this paper, this is done by introducing a methodology for learning the reduced data structure structures using reinforcement learning. The approach of a larger "teacher" network is the input and output of a "student" network compressed from the teacher network. In the first phase of this method, a day-to-day policy-maker network removes layers from a large "teacher" model. Secondly, a temporary policy grid again reduces the precision of the volume of each remaining layer. The result of the grid is to obtain a bonus-score based on the accuracy of network compression. The reward signal approach uses policy methods to educate a local optimal student network. Experiments show that compression can be achieved by more than 10x for models such as ResNet-34, while maintaining similar functionality to the Teacher's input network. It is also a valuable learning result that shows that policies that are trained in smaller "teacher" networks can be used to accelerate learning in larger "teacher" networks. [36]

The deep idea of Q-Learning is adapted to the field of continuous action. An actor-critic experimental algorithm is based on a definite policy gradient and can operate in continuous action spaces. Using the same learning algorithm, network architecture and the above parameters, the algorithm solves heavily more than 20 simulated physical tasks, including classical problems such as cartpole rotation, quick manipulation, foot motion, and vehicle driving. This algorithm is able to find policies whose performance is compatible with a programmed algorithm with full access to domain dynamics and derivatives. It is shown that for many tasks the algorithm can learn end-to-end lines: directly from raw pixels [38].

Now, CNN's design of the Neural Network (CNN) requires expertise and manpower. New architectures are built with accurate testing or a number of existing networks. MetaQNN is a robust learning-based modeling algorithm that automatically provides high-performance structures for CNN in a learning task. The learning agent continuously trains the CNN layers using Q-learning with Greedy's "ε" strategy and experience replay. Investigates a large but limited space of probable architectures and repeatedly finds plans with improved performance in learning. Based on image classification criteria, network-driven operating systems (only including standard canals, integrated circuits and fully connected layers), existing networks are designed to compete against advanced advanced techniques that are most used. With a variety of complex layers, existing modeling approaches improve network design in image classification tasks [43].

## 3. Cases and methods

Figure 1 shows an automatic compression agent, and the goal is to automatically calculate the sparsity ratio for each layer of the network in order to improve the quality of the compression algorithm. The sparse ratio can greatly affect the performance of the compression model [10,2]. The motivation for this task is for the sparsity ratio in precise details and with the learning of reinforcement that does not require human expertise. In particular, an agent receives and teaches a st embedding layer and obtains a precise sparse ratio for each layer. A state with accuracy in validating and without a precise reward setting is a very effective and efficient agent.

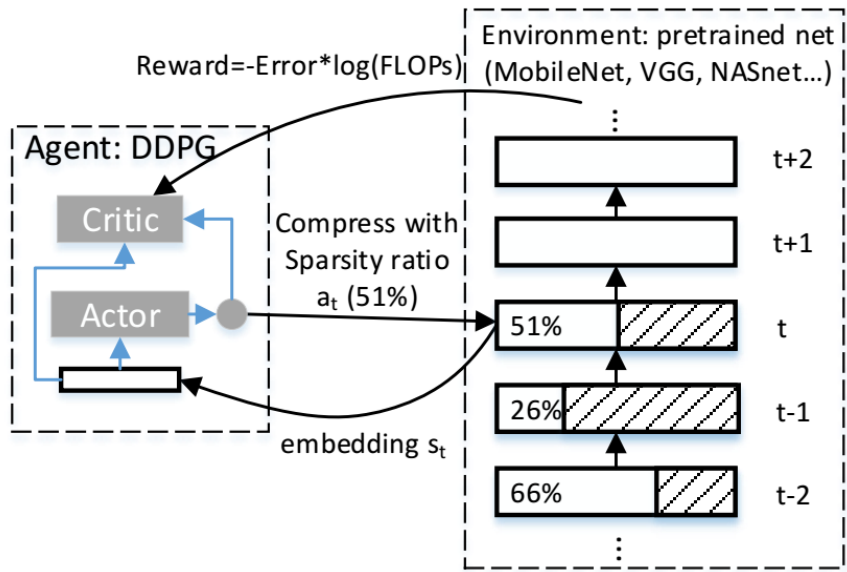

**figure 1.** A net of pre-training data that is processed in a layer in a layered manner. The agent receives an embedding for each T layer and generates a sparsity ratio. After compressing the layer, it moves to the next layer t + 1. After compressing all the layers, the approximate model is evaluated on valid images without accurate adjustment. Finally, the reward R is given to the agent [1].

Compression of the model is accompanied by a reduction in the number of channels from c to c' for each convolution layer, in which c '<c is obtained. Given the network algorithm and compression algorithm, the goal is precisely the effective number of c 'channels for each layer, which is usually determined in previous studies. [4, 37, 2]

Figure 2 shows the channel pruning algorithm for a single convolutional layer. We intend to reduce the width of the feature map B, and keep the exits in the C feature map. When the channels are cut, the channels can be removed from the filters that take these channels as inputs. Also, filters that generate these channels can also be deleted. It is clear that pruning the channel involves two main points. First, the channel selection is due to the need to select most of the show channels to maintain the same information. Second is reconstruction. The following maps must be reconstructed using selected channels [3].

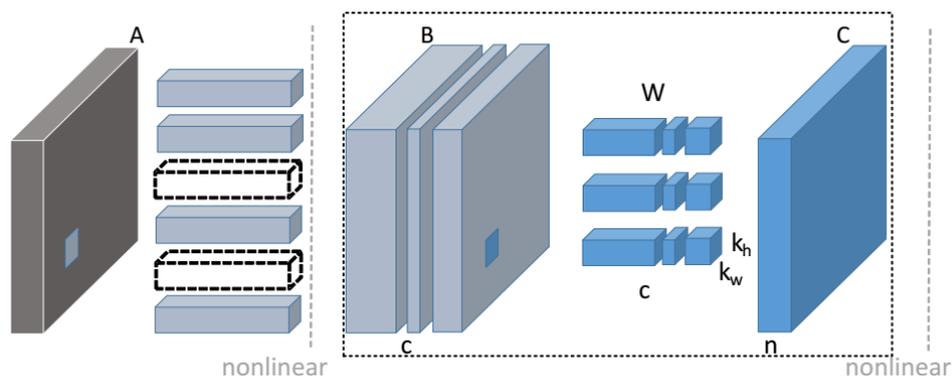

**figure 2.** Channel pruning to accelerate a convolutional layer. The intention is to reduce the width of the map to feature B, while the reconstruction error in the map is reduced. The optimization algorithm

is performed inside the dotted box, which does not include non-linear operation. Therefore, different channels of W filters can be deleted [3].

### 3.1. State Space

For each t layer, 11 attributes (which specify the state st state) are calculated:

$$(t, n, c, h, \omega, stride, k, FLOPs[t], reduced, rest, a_{t-1}) \quad (1)$$

The t id is the layer, the dimensions of the kernel n × c × k × k and the input c × h × w. The total number of reduced FLOPs in the previous layer's decreases. Rest is the number of remaining FLOPs in the previous layers. Before being transferred to the agent, it is between {1; 1}. Such features are essential for the agent.

The main lines of this design (Figure 3, middle) are mainly inspired by the philosophy of the VGG network [6] (Fig. 3, left). The cannulation layers are mostly 3 × 3 filters and follow two simple design rules: (1) For a size of output properties, the number of layers has the same number of filters. And (2) if the map size doubles, the number of filters doubles to maintain the complexity of time in each layer. It should be noted that this model, filters and complexity are less than the VGG network [6] (Fig. 3, left). The 34-layer base line has 3.6 billion FLOP, which is only 18% of VGG-19 (19.6 billion FLOP).

Based on a high-pitched network, shortcut connections (Figure 3, right) are inserted and transforms the network into a peer-to-peer version. When the dimension's increase (dotted line shortcuts in Figure 3), they consider two options: (A) The shortcut still performs identity mapping (with an extra zero for increased dimensions) and this option shows no additional parameters. (B) performs the prediction shortcut to match the dimensions (with 1 × 1 convolution) [1].

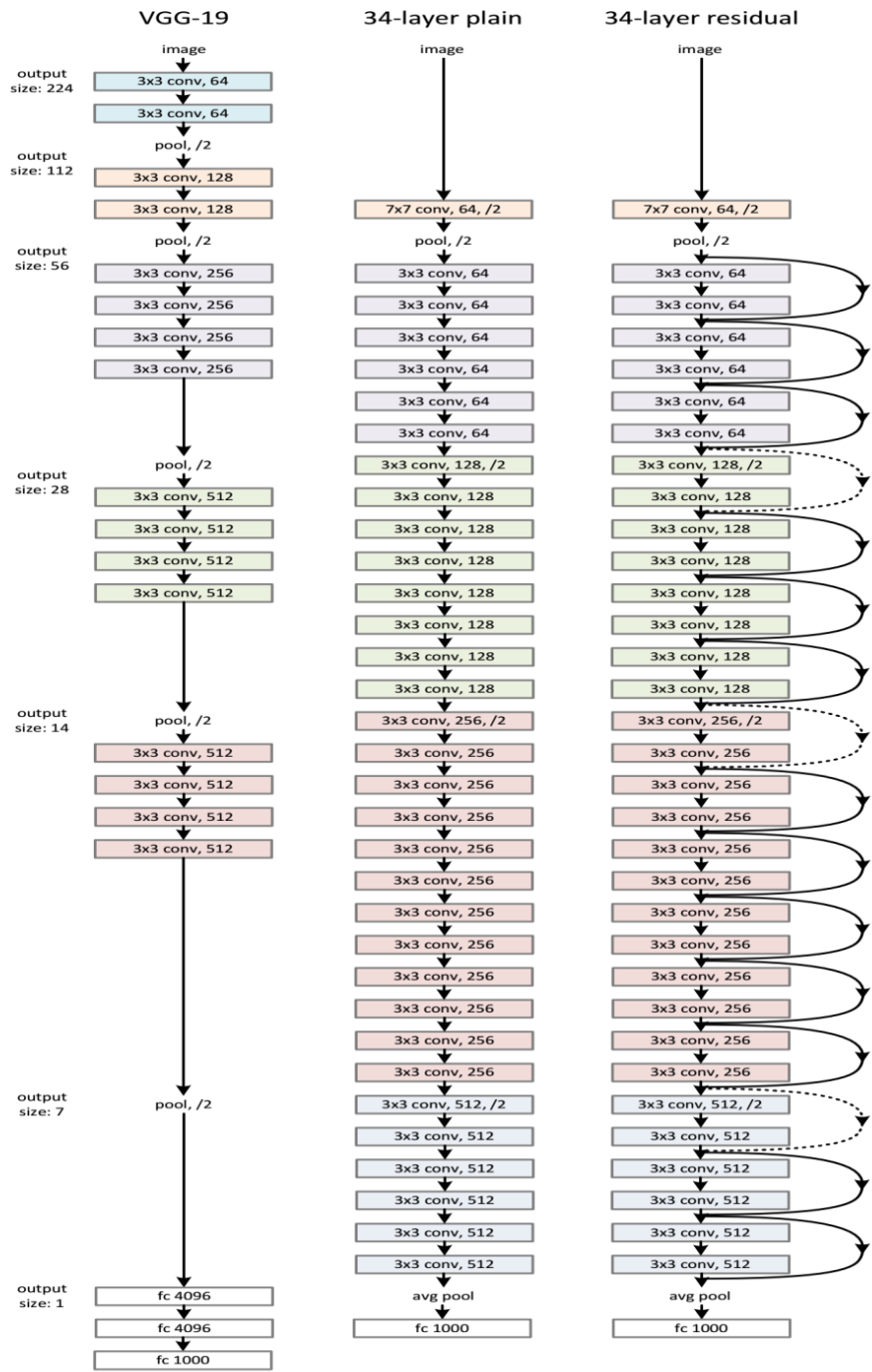

**Figure 3.** An example for the ImageNet network architecture. Left: VGG-19 [6] (19.6 billion FLOP) for reference. Mid dle: A simple network with 34 parameters (3.6 billion FLOPs). Right: A network with 34 parameters (3.6 billion FLOPs). Dotted shortcuts increase the dimension [1].

### 3.2. Action Space

In most of the existing works, the discrete space is used as an action space (such as {64; 128; 256; 512} for a number of channels). The action space may not have a high-precision model architecture search. However, it has been observed that compression is a highly sensitive spasticity model and requires a precise action space that results in a large number of discrete actions [1]. Such action spaces are very difficult for effective discovery [38]. Separation also reduces the order that it is suggested to use the continuous action space

$a = \frac{c'}{c}, a \in (0,1]$ ‹Which makes the compression possible more accurate.

By limiting the space of action (the sparsity ratio for each layer), you can correctly find the target compression ratio. [65, 5, 62] is used for the following reward:

$$R_{err} = -Error \qquad (2)$$

Obviously, this reward does not provide any incentive to reduce FLOP, so it should limit the scope of the action, noting that the algorithm is not limited to limiting FLOPs. FLOPs can be replaced with other resources, such as a number of parameters or even the actual conclusion time on a mobile device [1]. According to the tests, since the agent has no incentive to lower the budget, it can accurately reach the target's compression rate.

Agent is limited to specific actions to limit both state-action space and acceptable learning. First, it allows the agent to terminate the path at any point, for example, it may choose a termination state from the non-terminated state. In addition, it is only allowed to transfer a state with depth of layer i to a state with depth of the i + 1 layer to ensure that there are no loops in the graph [43].

### 3.3. Guaranteed compression accuracy

By fixing the reward function, you can precisely determine the amount of compression to be the most accurate. Experimentally, the error is proportional to the inverse (FLOPs) log. [39] Thus, the reward function is designed as follows:

$$R_{FLOP_s} = -Error . \log(FLOP_s) \qquad (3)$$

The reward function is error-sensitive; it also provides a small incentive to reduce FLOP. Based on the tests, the agent automatically determines the amount of compression.

### 3.4. Agent

As shown in Figure 1, the agent embeds the t layer from state st from the environment, and then receives the ratio of sparsity to action. The underlying layer is compressed using a compression algorithm (for example, spatial decomposition [4]) with ai. Then the agent moves to the next t + 1 layer and receives state st + 1. After completing the final layer T, the reward accuracy is evaluated on the validation set and returned to the agent. For quick exploration, bonus rewards are evaluated with no precise adjustment, and it works well.

The Deep Definitive Gradient Policy (DDPG) [38] is used for continuous control. The normal short distribution (TN) [40] is added to the policy μ for exploration of the μ' policy:

$$\mu'(\ ) \sim TN\big(\mu(S_t|\theta_t^\mu), \sigma^2, 0, 1\big) \qquad (4)$$

σ. $\mu'(s_t) \in (0,1)$ determines the randomness of the policy.

Following Block-QNN, [41] which uses various forms of the Bellmans equation, [42] each transition is in one part (st; at; R; st + 1), which is reward R after compression of the network. While updating, the base function b is included to reduce the gradient estimation variance, which is a pre-animated moving average. [42,34]:

$$Loss = \frac{1}{N}\sum_i (y_i - Q(s_i, a_i|\theta^Q))^2 \qquad y_i = r_i - b + \gamma Q(s_{i+1}, \mu(s_{i+1})|\theta^Q) \qquad (5)$$

The discount factor γ is set to 1 to not prioritize short-term rewards [43].

Q-Learning, the introduction of nonlinear approachers means that convergence is no longer guaranteed. However, such a cursory approximation is necessary for learning and generalization in large spaces. The NFQCA uses DPG update rules, but with the neural network performance approximation, it uses batch training for stability that can be solved with large networks. The contribution is to bring changes to DPG, inspired by the success of DQN, which allows it to use the neural network approximation to learn in large and practical online spaces. We refer to our algorithm as Deep DPG (DDPG, Algorithm 1) [38].

### Algorithm 1. DDPG algorithm

Randomly initialize critic network $Q(s, a|\theta^Q)$ and actor $\mu(s|\theta^\mu)$ with weights $\theta^Q$ and $\theta^\mu$.
Initialize target network $Q'$ and $\mu'$ with weights $\theta^{Q'} \leftarrow \theta^Q$, $\theta^{\mu'} \leftarrow \theta^\mu$
Initialize replay buffer $R$
**for** episode = 1, M **do**
   Initialize a random process $\mathcal{N}$ for action exploration
   Receive initial observation state $s_1$
   **for** t = 1, T **do**
     Select action $a_t = \mu(s_t|\theta^\mu) + \mathcal{N}_t$ according to the current policy and exploration noise
     Execute action $a_t$ and observe reward $r_t$ and observe new state $s_{t+1}$
     Store transition $(s_t, a_t, r_t, s_{t+1})$ in $R$
     Sample a random minibatch of $N$ transitions $(s_i, a_i, r_i, s_{i+1})$ from $R$
     Set $y_i = r_i + \gamma Q'(s_{i+1}, \mu'(s_{i+1}|\theta^{\mu'})|\theta^{Q'})$
     Update critic by minimizing the loss: $L = \frac{1}{N}\sum_i (y_i - Q(s_i, a_i|\theta^Q))^2$
     Update the actor policy using the sampled policy gradient:

## 4. Discussion and experimental results

For spatial decomposition [4], independent data retrieval is used (compression only works, weights and characteristics are not involved [44]). For the pruning channel [3], the maximum selectable response is used, and (pruning the weight according to size [12]) and the group of normal-coated layers [45] is maintained during the pruning, instead of being combined with the cannulation layers. The agent first examines 100 episodes with a constant noise of 0.5, and then extracts 300 episodes with a discontinuous exponential noise σ. The learning rate of 1e-4 and 10/1 is the main number of repetitions used for fine tuning [3].

### 4.1. dataset

4.1.1. CIFAR-10 dataset:

The CIFAR-10 dataset consists of 6,000 32x32 color images in 10 classes, with 6,000 images in each class, 50,000 educational pictures, and 10,000 experimental images. The datasets are divided into five instructional categories and one instructional class, each of which has 10,000 images. The test set contains exactly 1000 random images selected from each class. The instruction book contains the remaining images in random order, but some educational categories may include more images from a class. Between them, the tutorials include exactly 5000 photos from each class.

4.1.2. ImageNet Dataset:

ImageNet is an image database based on the WordNet hierarchy (currently only nouns) and displayed on each node by hierarchical representations of hundreds and thousands of images. There are currently more than 500 images per node on average. ImageNet is a useful resource for researchers, instructors, students, and more.

### 4.2. Compare with random search

Theoretically, with a large number of tests, random search results will be closer to the learning outcomes. It is unclear how many trials for success in random search are required, and of course for small networks. As shown in Figure 4, policy sparsity is searched for a space sub-division of 2 times. Before the 100th episode, the ADC behaves like a random search. Both are examined with a roughly constant oscillation. After the 100th episode, the ADC starts learning with exponential noise σ and can quickly find a good model; however, scattering is random. Exploring 400 episodes is not enough to search for random results and get results even for a small network in CIFAR-10.

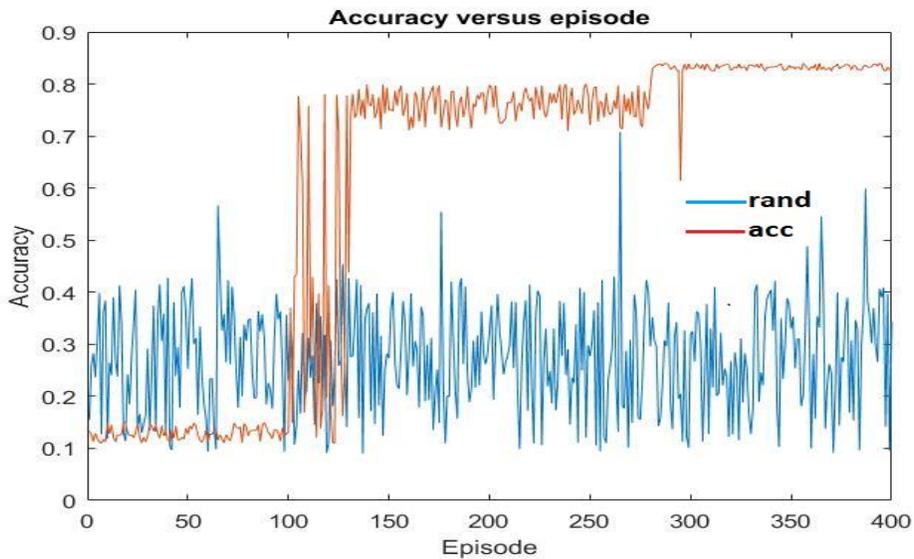

**Figure 4.** Validation accuracy against the episode, using spatial decomposition on CIFAR-10. Note that the best accuracy of random exploration occurs over a long period of time. A random search can not provide good results with limited parts.

### 4.3. FLOPs Limited Compression

This approach is compared with the three empirical policies [37, 3] in Figure 5: assemblies of uniform compression ratio - surface and depth - arrangement of surface and deep layers. Based on the distribution of the sparsity of different networks, a different strategy may be selected (eg VGG-16 superficial [3]). Empirical politics with a large margin is given preference. On the contrary, the best pruning regulated by the approach is very different from the experience (Fig. 5). This bottleneck architecture [5] is similar to a very dense architecture.

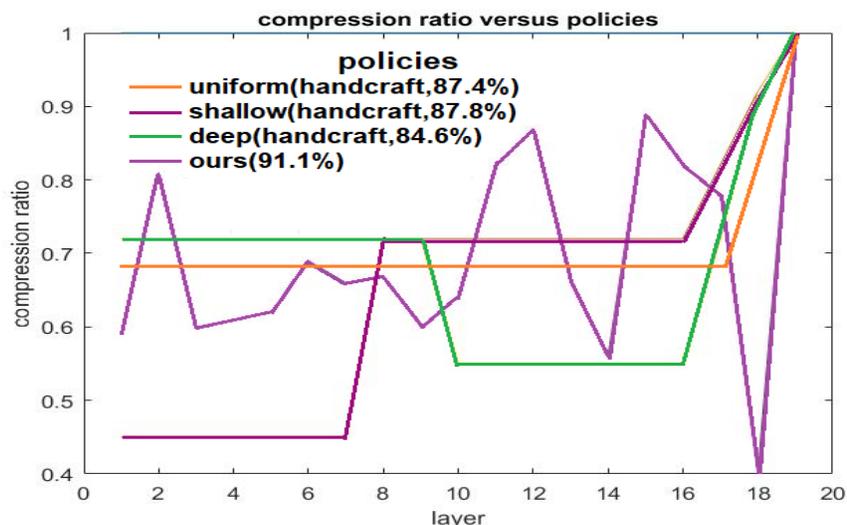

**Figure 5.** Comparison of pruning strategies is less than 2 times. A uniform policy collection is checked uniformly with the same compression ratio for each layer. Offensive pruning is superficial and deep-seated policies, and from the surface layers to the deep, respectively. The handy policy works best with the big margin. (It should be seen in color).

### 4.4. Guaranteed precision compression

Using reward, the agent can automatically compress compression by decreasing performance. The result has a compressive ratio of 91% with low degradation. The reward curve is obtained using reward Rerr. Note that if the compression is below 80%, the reward decreases sharply. Because RFLOP's reward focuses on error and at the same time provides less incentive for compression, the model is preferred with innocuous compression function. For excessive redundant networks, such as VGG-16 in ImageNet, RFLOP reward automatically maintains 65% FLOP without any damage. To shorten the search time, gain reward with precision validation and without accurate adjustment. By fine-tuning the reward, the agent becomes heavily compressed, with the exact adjustment accuracy close to the original precision.

### 4.5. Generic ability

In order to evaluate the generalizability of the compact model for learning to transfer the object-recognition work, the VGG-16 compression test is tested [50] (4-fold reduction of the FLOP, 1.2% increase in error by channel decompression). Simply use VGG-16 as a column for Faster R-CNN [51] and it is published and tested with the same settings. The performance is evaluated with an average mAA (mAP) and mAP (initial measurement of COCO competition [39]).

It is shown in Table 1, although the compressed model is only 0.6% better than the ImageNet classification, [3], however, the boundary of object detection is increased (2.1% mAP improvement [0.5.0.95]). The ADC is ahead of the 0.5% line. This significant increase is assumed to be due to the precision compression policy and compact architecture that is well-tuned.

**Table 1.** Compare VGG-16 compression for Faster R-CNN detection. In accordance with the diagnostic function, it is the best compression model produced by the ADC that provides better performance in dimensional detection.

|  | mAP (%) | mAP [.5, .95] (%) |
|---|---|---|
| baseline | 68.7 | 36.7 |
| 2× handcrafted [3] | 68.9 (-0.6) | 36.7 (-0.0) |
| 4× handcrafted [3] | 67.3 (-1.8) | 35.1 (-1.4) |
| 4× handcrafted [8] | 66.8 (-0.9) | 36.5 (-0.2) |
| 4× ADC (ours) | **68.8 (+0.1)** | **37.2 (+0.5)** |

### 4.6. Dense network pruning

Recently, fast and smaller dense networks [52, 7, 53] have been designed for ever-increasing needs. The focus of this article is on ImageNet, a very dense network that includes deep convolution and point-to-point convolution. Previous attempts to use the manual policy for ImageNet resulted in degradation of accuracy [37]. ImageNet's version is up to 75.5% of the original parameters at 67.2% accuracy, which is even worse than 0.75 MobileNets (61.9% with a top-1 accuracy of 68.4%). However, ADC pruning policy significantly improved the quality of pruning: ADC-pruned MobileNet 68.8% of Top-1 accuracy with 285 FLOP compared to the original 0.75 MobileNet 68.4% Top-1 accuracy with 325 FLOP, was evaluated in all ImageNet.

As shown in Figure 6, a shallow policy is slightly worse than the original ImageNet under the 2-fold reduction of the FLOP. ADC pruning policy improves ImageNet model pruning. To reduce the 2-fold of the FLOPs, it reached 68.8% of Top-1 accuracy, and 67.5% improvement in empirical policy with reward RFLOPs, calculating 14% without destruction.

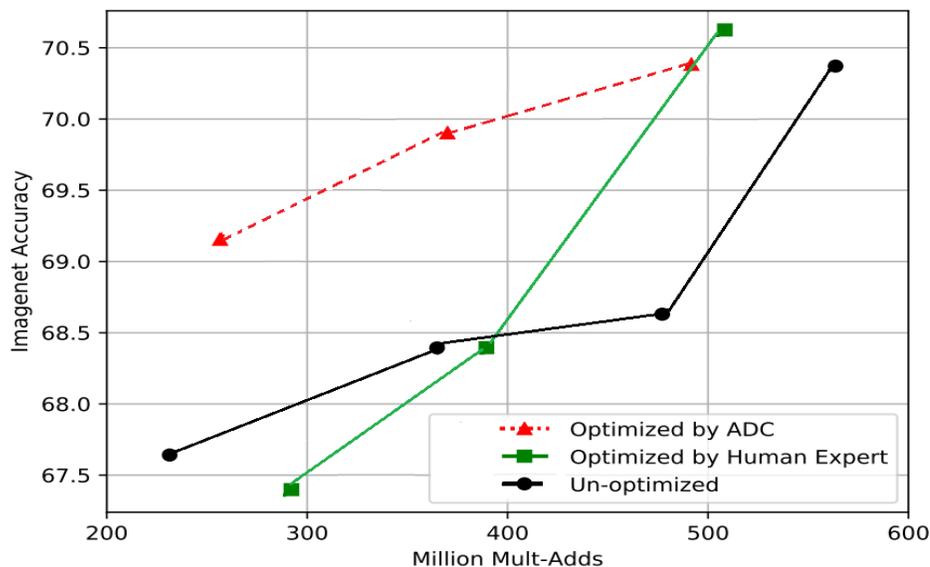

**Figure 6.** Original Pruning ImageNet is shown as black links. The result is a worse and less shallow policy than the original ImageNet. Due to ADC policies, channel pruning will greatly increase MobileNets performance. RFLOPs reward naturally to 86% of non-destructive FLOPs. (It should be seen in color).

The model is tested on an Android phone. 1.87 times the speed for convolutional layers of $1 \times 1$ and 1.65 times the speed, in general, under a 2-fold reduction of the FLOP. There is not much real time to accelerate deep convolution. Note that although deep convolution layers provide much less computation, it still consumes a lot of time. Effective implementation of deep convolution is difficult due to the small proportion of computing for fetch memory. Compact models also consume less memory (VSS, virtual volume volume). For manual policies, it's hard to get speed. Accurate ADC policies are essential to success. It has also been shown that the ADC continues to operate on smaller neural networks and has an architecture that is found by learning enhancements such as NASNet [32].

## 5. Conclusion

Conventional compression techniques use manual features and require specialist expertise to learn large design space and measure volume, speed, and accuracy, which are usually efficient and cost-effective. In this paper, it has been suggested that automated deep compression (ADC), which is automatically enhanced with learning learning in the design environment, significantly improves the compression quality of the model. Two reward schemes are limited to both compression and compression is guaranteed and carefully designed. Significant results for CIFAR-10, ImageNet are provided. The compact model defines the object as accurately as PASCAL VOC. The ADC technique facilitates a deep perspective on mobile devices.